\documentclass[conference]{IEEEtran}
\IEEEoverridecommandlockouts
\usepackage{cite}
\usepackage{amsmath,amssymb,amsfonts}
\usepackage{algorithmic}
\usepackage{graphicx}
\usepackage{textcomp}
\usepackage{xcolor}
\usepackage{subcaption}
\usepackage{pgfplots}
\usepackage[vlined]{algorithm2e}

\usepackage[T1]{fontenc}
\usepackage[acronym,shortcuts,smallcaps]{glossaries}

\renewcommand{\ac}[1]{\gls{#1}}
\renewcommand{\acp}[1]{\glspl{#1}}

\def\BibTeX{{\rm B\kern-.05em{\sc i\kern-.025em b}\kern-.08em
    T\kern-.1667em\lower.7ex\hbox{E}\kern-.125emX}}

\newglossaryentry{MDLSTM}
{
  name={MDLSTM},
  description={multi-dimensional long short-term memory},
  first={\glsentrydesc{MDLSTM} (\glsentrytext{MDLSTM})},
  plural={MDLSTMs},
  descriptionplural={multi-dimensional long short-term memories},
  firstplural={\glsentrydescplural{MDLSTM} (\glsentryplural{MDLSTM})}
} 

\newglossaryentry{LSTM}
{
  name={LSTM},
  description={long short-term memory},
  first={\glsentrydesc{LSTM} (\glsentrytext{LSTM})},
  plural={LSTMs},
  descriptionplural={long short-term memories},
  firstplural={\glsentrydescplural{LSTM} (\glsentryplural{LSTM})}
} 

\newglossaryentry{RNN}
{
  name={MDLSTM},
  description={recurrent neural network},
  first={\glsentrydesc{RNN} (\glsentrytext{RNN})},
  plural={RNNs},
  descriptionplural={recurrent neural networks},
  firstplural={\glsentrydescplural{RNN} (\glsentryplural{RNN})}
} 

\newglossaryentry{CER}
{
  name={CER},
  description={character error rate},
  first={\glsentrydesc{CER} (\glsentrytext{CER})},
} 

\newglossaryentry{WER}
{
  name={WER},
  description={word error rate},
  first={\glsentrydesc{WER} (\glsentrytext{WER})},
} 

\newglossaryentry{LMBR}
{
  name={LMBR},
  description={last-minute batch-padding},
  first={\glsentrydesc{LMBR} (\glsentrytext{LMBR})},
} 

\newglossaryentry{HTR}
{
  name={HTR},
  description={handwritten text recognition},
  first={\glsentrydesc{HTR} (\glsentrytext{HTR})},
} 

\newglossaryentry{HR}
{
  name={HR},
  description={handwriting recognition},
  first={\glsentrydesc{HR} (\glsentrytext{HR})},
}

\newglossaryentry{NHR}
{
  name={NHR},
  description={neural handwriting recognition},
  first={neural handwriting recognition (\glsentrytext{NHR})},
} 

\newcommand{\EXCLUDE}[1]{}

\begin{document}

\title{No Padding Please:\\ Efficient Neural Handwriting Recognition} 

\author{\IEEEauthorblockN{1\textsuperscript{st} Gideon Maillette de Buy Wenniger}
\IEEEauthorblockA{\textit{The Adapt Centre}\\ \textit{Dublin City University} \\
Dublin, Ireland \\
gemdbw@gmail.com}
\and
\IEEEauthorblockN{2\textsuperscript{nd} Lambert Schomaker}

\IEEEauthorblockA{
\textit{Bernoulli Institute for Mathematics,}\\ 
\textit{Computer Science and}\\
\textit{Artificial Intelligence}\\
\textit{University of Groningen} \\
Groningen, The Netherlands \\
l.r.b.schomaker@rug.nl}
\and
\IEEEauthorblockN{3\textsuperscript{th} Andy Way}
\IEEEauthorblockA{\textit{The Adapt Centre}\\ \textit{Dublin City University} \\
Dublin, Ireland \\
andy.way@adaptcentre.ie}
}

\maketitle

\begin{abstract}
\Gls{NHR} is the recognition of handwritten text with deep learning models,
such as \ac{MDLSTM} recurrent neural networks. 
Models with \ac{MDLSTM} layers have achieved state-of-the art results on handwritten text recognition
tasks. While multi-directional \ac{MDLSTM}-layers have an unbeaten ability to capture the complete context in all directions, this strength 
limits the possibilities for parallelization, and therefore comes at a high computational cost.

In this work we develop methods to create efficient \ac{MDLSTM}-based models for \ac{NHR}, particularly a method 
aimed at eliminating computation waste that results from padding. This proposed method, called \emph{example-packing},
replaces wasteful stacking of padded examples with efficient tiling in a 2-dimensional grid.
For word-based \ac{NHR} this yields a speed improvement of factor 6.6 over an already efficient baseline of minimal 
padding for each batch separately. For line-based \ac{NHR} the savings are more modest, but still significant. 

In addition to example-packing, we propose: 1) a technique to optimize parallelization for dynamic graph definition 
frameworks including PyTorch, using convolutions with grouping, 2) a method for parallelization across GPUs for variable-length
example batches.
All our techniques are thoroughly tested on our own PyTorch re-implementation of \ac{MDLSTM}-based \ac{NHR}
models. A thorough evaluation on the IAM dataset shows that our models are performing similar to earlier implementations of state-of-the art models. 
Our efficient \ac{NHR} model and some of the reusable techniques discussed with it offer ways to realize relatively efficient models 
for the omnipresent scenario of variable-length inputs in deep learning.

\end{abstract}

\begin{IEEEkeywords}
variable length input, example-packing, multi-dimensional long short-term memory, handwriting recognition, deep learning, fast deep learning
\end{IEEEkeywords}

\section{Introduction}

Over the last few years, end-to-end deep learning models for automatic handwriting recognition \cite{Bluche2013FeatureEW,PhamEtAl2014,Voigtlaender2016} have started to become competitive with earlier approaches, such as those based on hidden Markov models \cite{MartiAndBunke:2001, BengioEtAl2004}.\footnote{Sometimes handwriting recognition is  called \ac{HTR} in the literature. We opted for ``handwriting recognition'' for brevity, and also taking into account a markedly higher google n-gram frequency.}
A key ingredient to the success of these models has been the application of \acp{MDLSTM} \cite{mdlstmsGraves2007}.
However, the successful application of \acp{MDLSTM} for \ac{HR} is complicated by two main factors: 1) their computational 
cost and 2) their instability during learning. Because of these challenges, some researchers have questioned the need for using 
\acp{MDLSTM} in the first place \cite{Puigcerver2017}, and/or suggested to (partly) replace them by convolutional layers which are better understood 
and easier to use out of the box in most deep learning frameworks \cite{CarbonellEtAl2018}.
But despite the difficulties, \acp{MDLSTM} and variants have a strong theoretical strength, 
which is their ability to capture the complete surrounding context at every cell in an \ac{MDLSTM}-layer. This is particularly true 
for the multi-directional version of \acp{MDLSTM}, for example the 4-directional \ac{MDLSTM} combines the complete context of all 
cells around a particular cell, in each of the four scanning directions. Whereas deep convolutional networks might be argued to be able
to approximate this, the conceptual elegance of \acp{MDLSTM} combined with their empirical success for \ac{HR}
as described in recent literature \cite{PhamEtAl2014,Voigtlaender2016} make it unattractive to dismiss them altogether. 
As such, this paper focusses on the question: when MDLSTMs are applied for neural \ac{HR}, 
how can this be done effectively and efficiently?

Concerning computational cost, a major problem with the original implementations of \acp{MDLSTM} was that 
the inherent sequential dependencies in the computation forced these implementations to compute each layer cell-by-cell.
A big leap has been made by the insight that in fact the dependencies in \acp{MDLSTM} computation still allow for 
sets of cells to be computed in parallel, conceptually scanning an image diagonally in parallel, rather than column by column 
as in the naive earlier implementations. Furthermore, using a smart reorganization 
of the input, which we will refer to in this paper as the \emph{input-skewing trick} \cite{VanDenOord:2016:PixelRecurrentNeuralNetworks}, this parallel computation can be done efficiently 
on GPUs using convolution as a the workhorse. 

Concerning computational stability, it has been found that \acp{MDLSTM} exhibit a severe conceptual problem in that the values of the memory-state tensor of 
\acp{MDLSTM} can grow drastically over time and even become infinite. This can hamper learning, or even derail it altogether. 
This phenomenon and the mathematical principles behind it have been thoroughly described by \cite{LeifertEtAl2014}. More importantly, 
the authors also describe improved versions of \acp{MDLSTM}, the most effective of which is called \emph{Leaky-LP cell}. These \emph{stable cells} overcome 
the instability problems of \acp{MDLSTM} while retaining their key ability to preserve state, i.e. keep things in memory, over a long time. The application of 
these stable cells was found to be crucial by \cite{Voigtlaender2016}, and we share their finding when re-building models to reproduce state-of-the art 
literature results for neural \ac{HR}.

Inspired by the earlier work of  \cite{PhamEtAl2014}, \cite{Voigtlaender2016}
and \cite{VanDenOord:2016:PixelRecurrentNeuralNetworks}, we started out on this work with the conviction that it should be possible 
to build effective as well as computationally efficient \ac{MDLSTM}-based \ac{HR} models while sticking to existing deep learning frameworks. 
This paper is the result of this mission. It contributes 
to the field with a thorough discussion of what is needed to reproduce state-of-the art \ac{HR} results with \ac{MDLSTM}-based models. 
It introduces an array of techniques that can be applied to make \acp{MDLSTM} fast, while using standard deep learning frameworks. Finally, it proposes 
a new technique called \emph{example-packing} which can be used to eliminate the majority of padding when dealing with variable-sized inputs. This technique 
alone can yield major computational gains by drastically reducing the waste of GPU memory and computation spend on padding. The saved memory enables larger 
batch sizes, yielding significant speedups.

\begin{figure*}[h!]
\begin{center}
  \includegraphics[scale=0.20]{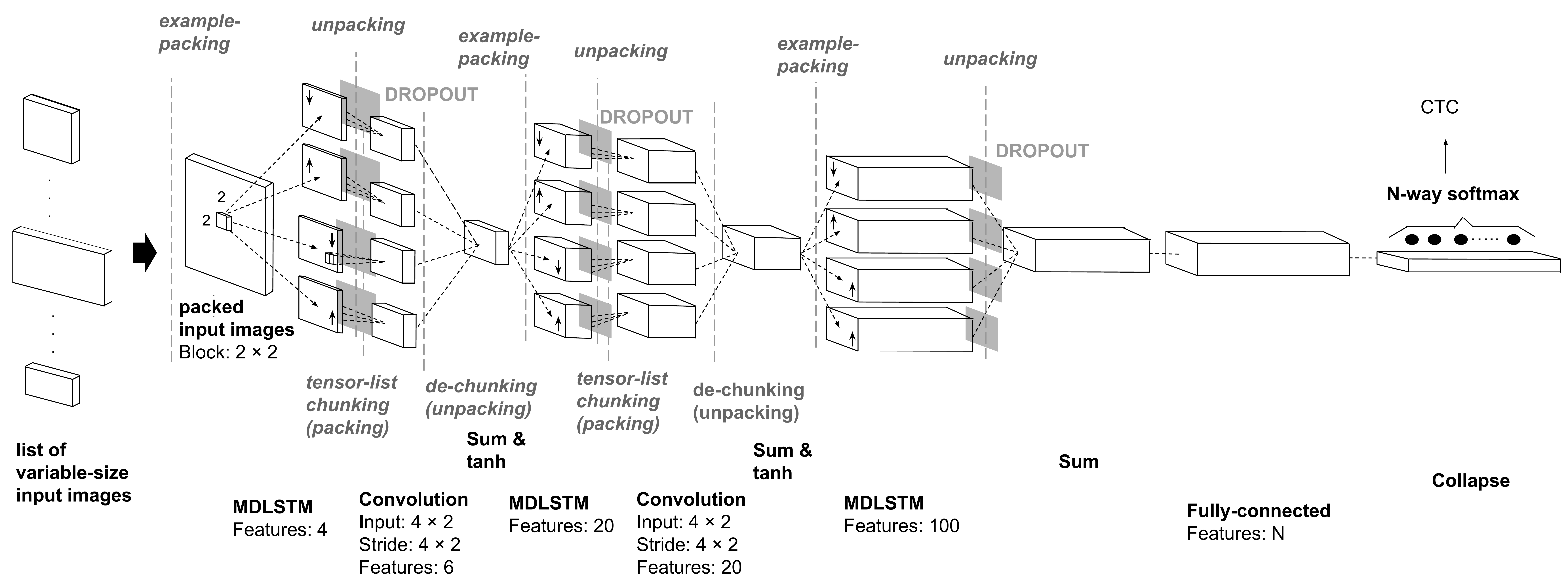}
\end{center}
\caption{Network model structure, adapted from \cite{PhamEtAl2014}, with places where packing/unpacking are applied for efficient computation.}
\label{figure:network-model-structure}
\end{figure*}

\section{Overview}
\glslocalreset{NHR}
The term handwriting recognition is sometimes used to cover the whole range of sub-tasks associated with handwritten text, 
including for example \emph{word spotting} \cite{RathAndManmatha2007, FischerEtAl:2012}, which groups word images into clusters of similar 
words. In the context of this publication however, we use \ac{NHR} to refer specifically to the task of creating text output from 
handwritten text images. More precisely: 1) the input is in the form of (cut-out) images of handwritten text, i.e. word-strips or line-strips, 2) output is in the form of character or word-sequences, 3) a deep learning model is used to produce sequences 
of character probabilities, this is used in combination with connectionist temporal classification (CTC) \cite{Graves:2006:CTC}
and the associated loss-function, 4) the whole system is trained end-to-end with mini-batches of labeled examples, using 
back-propagation and other standard deep learning techniques.
Our  model is chosen to be almost identical to the one used in \cite{PhamEtAl2014}, 
with the difference that we share the last, fully-connected layer across directions.
Figure \ref{figure:network-model-structure} shows the model structure.\footnote{Note that input/stride sizes, e.g.  $4 \times 2$, are of the form height $\times$ width in this diagram.}
It also includes the places where \emph{example-packing} can be applied,
to process variable-sized examples more efficiently using minimal padding. Example-packing is discussed in section \ref{section:example-packing}.
In Appendix C we discuss details regarding the software that has been used in this work.

\section{Leaky LP cells: a stable version of MDLSTMs}
\label{section:leaky-lp-cells}

\begin{figure*}[!htb]
\begin{center}
\begin{minipage}{.54\textwidth}
  \includegraphics[scale=0.30]{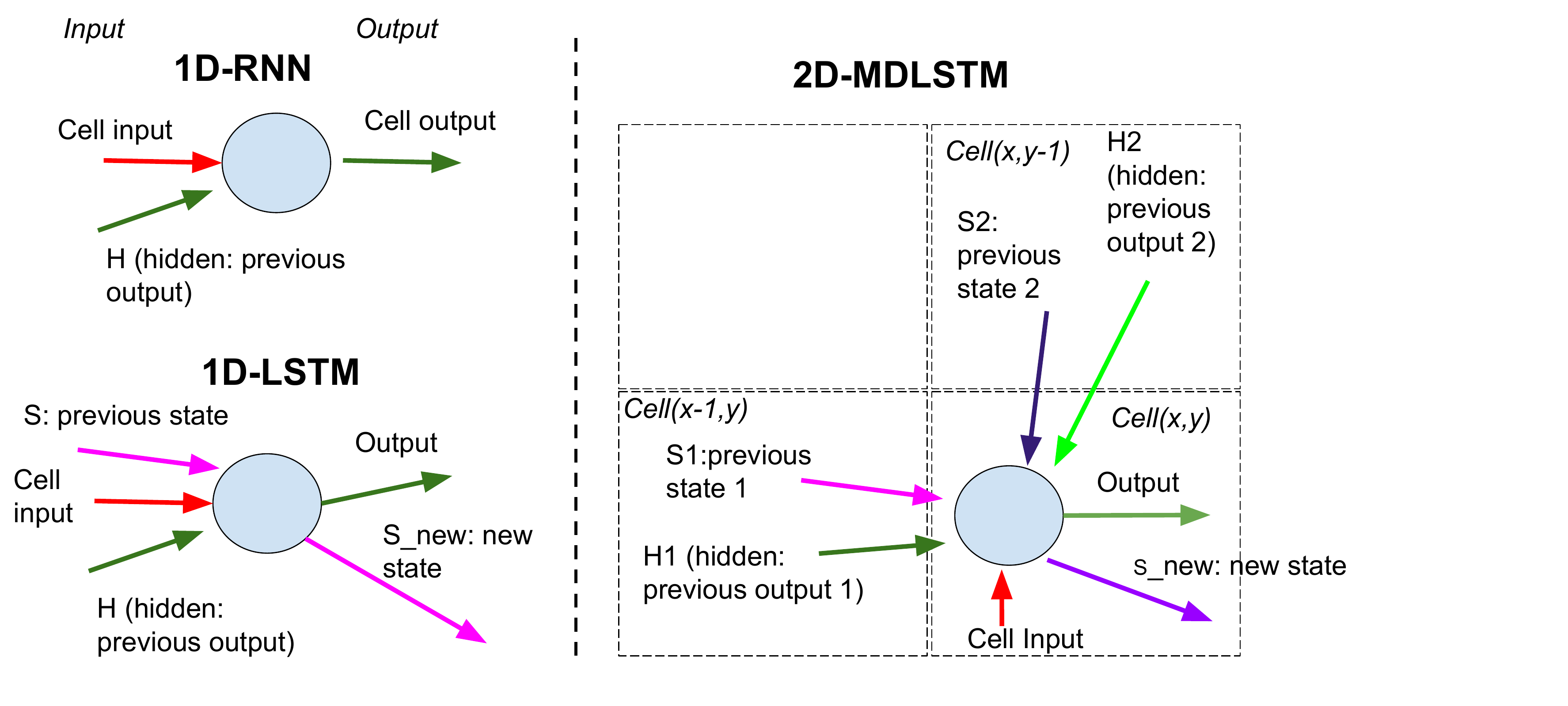}
 \caption{1-D LSTM versus 2-D-MDLSTM computational structure}
\label{figure:2dmdlstm-explained-from-1d}
\end{minipage}
\begin{minipage}{.42\textwidth}
  \includegraphics[scale=0.5]{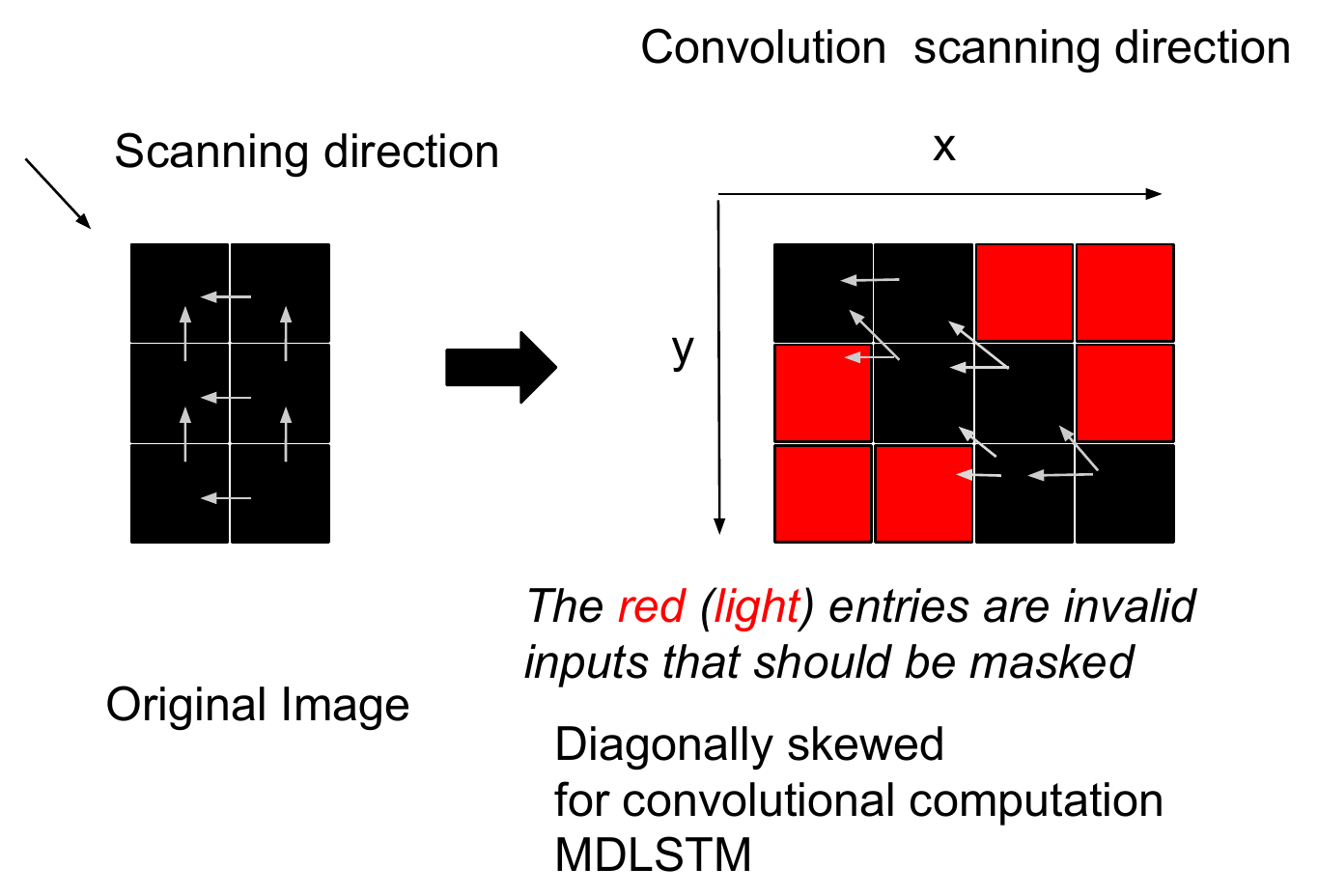}
 \caption{The \emph{input-skewing} trick.} 
 \label{figure:input-skewing-trick}
\end{minipage}
\end{center}
\end{figure*}

\begin{figure*}
\begin{center}
\hspace{-2cm}
\begin{subfigure}[b]{0.37\textwidth}
  \includegraphics[scale=0.3]{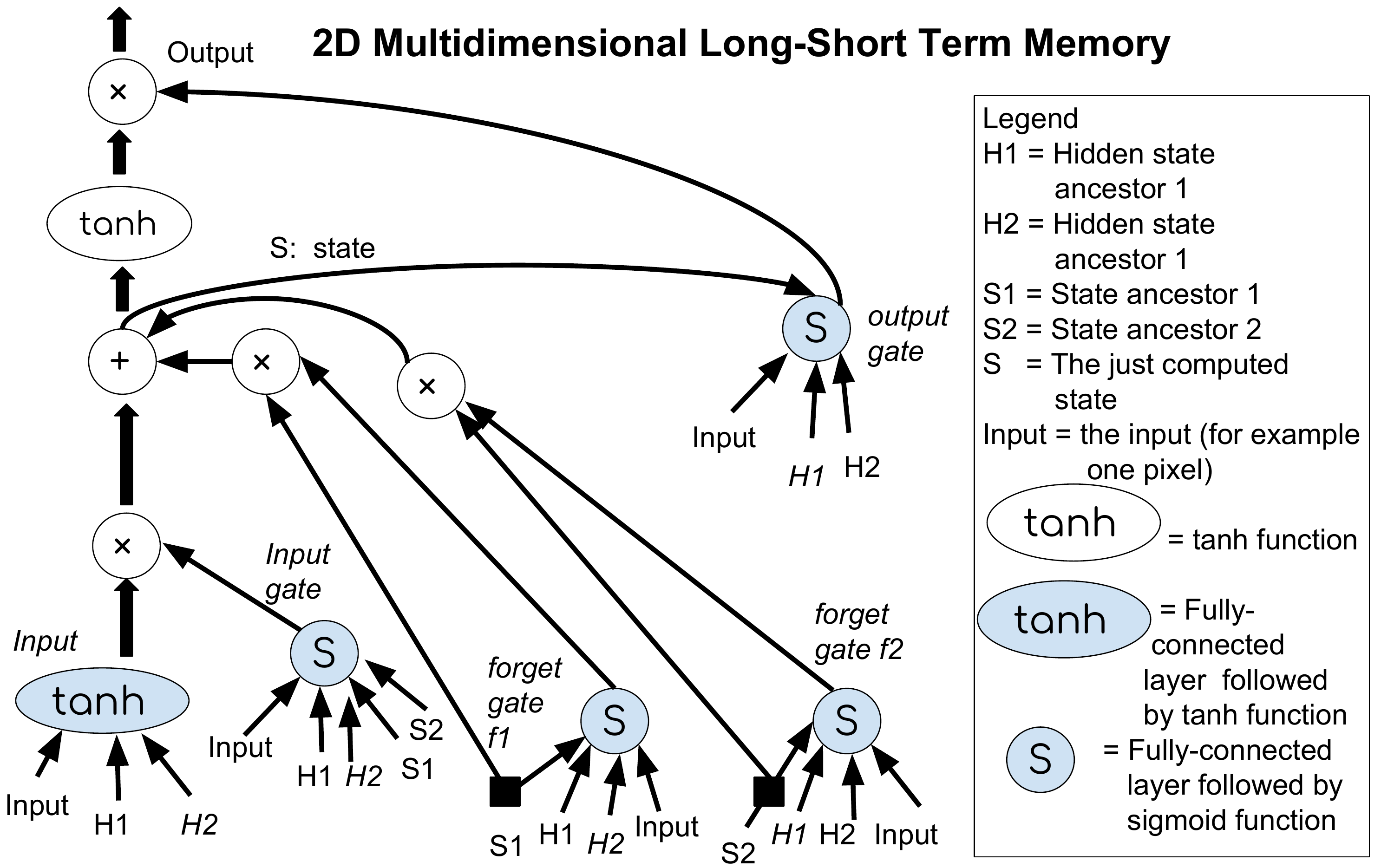}
 \caption{2-D MDLSTM computational graph.}
\label{figure:2dmdlstm-compuational-graph}
\end{subfigure}
\hspace{2.5cm}
\begin{subfigure}[b]{0.37\textwidth}
 \includegraphics[scale=0.3]{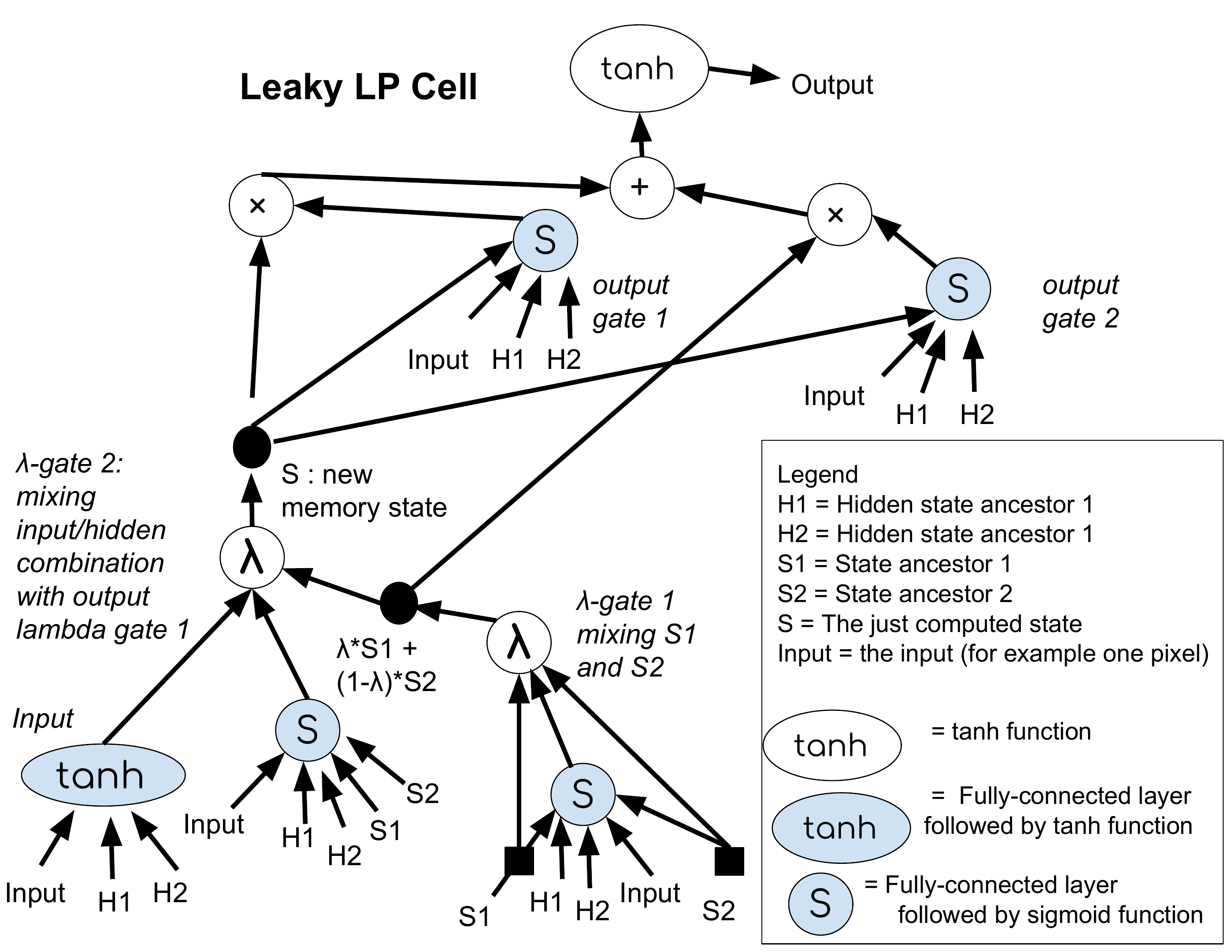}
 \caption{2D Leaky LP cell computational graph: with $\lambda$ gates.}
\label{figure:leaky-lp-cell-compuational-graph} 
\end{subfigure}
\caption{Computational graph for \ac{MDLSTM} and its stable variant Leaky LP cell.}
\end{center}
\end{figure*}

\acp{MDLSTM} \cite{mdlstmsGraves2007} are the multi-dimensional extension of \acp{LSTM}. 
In this paper we focus on the 2-dimensional version of \acp{MDLSTM}, which is the version 
used for \ac{NHR}. For readers unfamiliar with the details of this type of network cells, it is helpful 
to make a comparison with one-dimensional \acp{RNN} and \acp{LSTM} \cite{Hochreiter:1997:LSMT}, 
see Figure \ref{figure:2dmdlstm-explained-from-1d}.
For simple 1D-\acp{RNN} there is one input and hidden unit, and one output. For \acp{MDLSTM} this is extended with an additional 
input state and output state. In contrast, for 2-D-\acp{MDLSTM} there are two 
neighboring \emph{predecessor} cells in a conceptual 2D grid of cells used for computation. Each neighbor provides a pair of 
of $\langle$\emph{hidden}, \emph{state}$\rangle$ inputs, with the other input called \emph{cell input}   
coming from the computed cell, as in the 1D case. 
Furthermore,  stacking the output of four different \acp{MDLSTM}, one for each possible direction 
the 2-D grid can be scanned, yields a 4-directional 2-D-\ac{MDLSTM}. 
This version is the one typically used for \ac{NHR}.
Figure \ref{figure:2dmdlstm-compuational-graph} shows the computational graph for \acp{MDLSTM}. A crucial role is played 
by two forget gates used to weigh and then combine the states $S_1$ and $S_2$, obtained as inputs from the two predecessor cells. 
The value of these gates is in the range zero to one, and here lies a problem. When the sum of the gate activations becomes larger than one, 
the absolute values of the state entries, i.e. the norm of the state, can grow over time. As thoroughly discussed in \cite{LeifertEtAl2014}
and confirmed in our own experiments, this is a real problem and not just a theoretical one. States can grow rapidly over time, 
and even become infinity, and gradient clipping cannot fix this either. A naive solution would be to simply multiply 
the output of both gates by a factor 0.5, guaranteeing that the combined gate activation never exceeds 1. But \cite{LeifertEtAl2014}
note that while fixing the instability problem, this causes a new problem by making it impossible for the cell to preserve state, that is 
remember, over a long time. Therefore, they proposed a more rigorous solution, introducing so-called \emph{lambda-gates} that 
produce a weighted sum of the inputs, with the weights being predicted by the gate and summing to one, see Figure \ref{figure:leaky-lp-cell-compuational-graph}. 
This solves the severe problems of \ac{MDLSTM} instability, as also reported by \cite{Voigtlaender2016}, 
while retaining the crucial ability to preserve state over a long time. 

Based on the idea of lambda-gates, multiple variants of stable \ac{MDLSTM} cells are possible, as discussed in \cite{LeifertEtAl2014}, with the best performing 
one being the Leaky LP cell, but all of them yielding solid results. In this work, we use a slight variant of the Leaky LP cell: 
the previous memory state is used in place of the newly computed memory state as (memory) input to the two output gates. 
This variant yielded faster learning and superior results in our experiments.

\section{Parallelizing the computation}

To make  training and application of \ac{MDLSTM}-based \ac{NHR} models computationally feasible, 
parallelization is crucial. 
 Parallelization over the batch computation is the simplest form of parallelization, and is 
available without any additional effort in most deep learning frameworks. 
However, due to the large amount of memory required by each example and its \ac{MDLSTM} representations, 
it is typically not possible to drastically increase the batch size as a simple way to increase parallelization.
The exact batch size possible for \ac{NHR} will depend per problem and setup\footnote{Being determined by GPU memory, height, width, number of (color) channels of the examples, network structure and parameterization and other factors.},
but in practice very large batch sizes are not possible. 
For this reason, additional other forms of parallelization should be considered and implemented wherever possible.

\subsection{Parallel column computation using the input-skewing trick}
The next optimization with large impact is the parallelization of \ac{MDLSTM} computation within image columns, introduced by \cite{VanDenOord:2016:PixelRecurrentNeuralNetworks},
increasing the level of parallelism by a factor as large as the example height. Notably, a similar trick was discovered
independently by \cite{Voigtlaender2016}, but their variant relied on low-level implementation rather than leveraging existing tools, particularly efficient implementations of convolution, within deep learning frameworks.
As explained in section \ref{section:leaky-lp-cells}, computation of 2-D \acp{MDLSTM} can be modeled as a grid of computational ``cells'', 
whereby each cell takes context input on two neighboring ancestor cells, one above and one on the left (as defined by the scanning direction). 
Figure \ref{figure:input-skewing-trick} shows what we will call the \emph{input-skewing} trick: 
each of the $n-1$ image rows $r_i, i \in [2,n]$ is shifted  
with an increasing number of $i-1$ pixels. The results is a diagonally skewed input image that can be  
applied for efficient computation of the \ac{MDLSTM} using convolution. 
The input-skewing trick transforms the original input into a row-shifted version. This produces a conceptual grid of computational cells,
whereby each column of cells depends only on cells in the previous column. 
The skewed input image and corresponding computational grid, contains cells that correspond to valid/invalid inputs, 
marked by black/red squares in Figure \ref{figure:input-skewing-trick}. 
A binary mask tensor with ones/zeros for the valid/invalid cells is used to mask invalid inputs during computation.
Skewed input image plus mask thereby enable fast computation with framework-native convolutional layers, 
without low-level programming.

\subsection{Convolutions with Grouping}

Another technique to further increase parallelism, 
relevant to deep learning frameworks with \emph{dynamic graph definition} including PyTorch, 
is to make use of \emph{convolutions with grouping}.\footnote{The technique may not be relevant for tensorflow and other frameworks working with static computational graph definition. At the expense of less flexibility, the pre-compilation and optimization of the computational graph in such frameworks solve some of
the problems of poor automatic parallelization that occur when working with dynamic computational graphs.}
While conceivably this technique has been used in other domains, to the best of our knowledge, it has not been proposed for \ac{NHR}.
In convolution with grouping, the computation is partitioned into $m$ input groups and $n$ 
output groups. Each of the $n$ outputs is only connected to the nodes of one of the $m$ input groups, whereby $n$ must be an exact multiple of $m$.
The output of a convolution network with $m$ input and $n$ output groups is the same as for $m \times n$ separate networks, each with $\frac{1}{m}$th of the inputs connected to one $\frac{1}{n}$th of the outputs; but the difference is that these $m \times n$ grouped computations are performed in parallel rather than sequentially.

Figure \ref{figure:2dmdlstm-compuational-graph} shows the computational graph for 2D-\ac{MDLSTM} computation. 
Taking into account the structure of the graph, there are three main observations that enable 
optimization of \ac{MDLSTM} computation (and analogously Leaky LP cell computation), using convolutions with grouping: 
\begin{enumerate}
\item The computation of one MDLSTM cell / a column of MDLSTM cells (using the input-skewing trick), can be divided into 
a set of operations that are mutually independent, and hence can be computed in parallel. 
\item Most of the computations rely on the current pixel input and/or the ancestor hidden and memory states as inputs. 
\item 
Even when the inputs differ, it remains possible to parallelize multiple computations using a single convolution with grouping.
This is done by having multiple input groups in addition to multiple output groups.
Finally, in cases where the number of outputs per input differs per input, 
this can be solved by replicating some of the inputs multiple times. This gets around the typical restriction that $n$ must 
be an exact multiple of $m$. 
\end{enumerate}

In the concrete case of 2D-\ac{MDLSTM} computation (see Figure \ref{figure:2dmdlstm-compuational-graph}), there are five matrix computations necessary to get input activations for the 
input node, input gate, two forget gates, and output gate. 
Since these input activation computations depend only on
the input, either an image or input from the previous network layer, they can all be computed in parallel using a 2-dimensional
convolution. This convolution is of size $1 \times 1$ for both filters and stride, to avoid overlap between convolution kernel applications.
It is then further parallelized using grouping, in this case with just one input group and five output groups.
Convolutions with grouping are similarly applied to parallelly compute the multiple tensors 
that use the same hidden or memory states as the input. Details are given in Appendix A.

\subsection{Example-list based multiple-GPU training}
Parallelization over multiple GPUs is another way to further increase the speed of computation.
This type of parallelization is typically supported natively in deep learning frameworks, for example 
in PyTorch there is a method \emph{DataParallel} that supports it. However, the problem with DataParallel 
is that it only works with tensors of the same size. Essentially it expects a data and label tensor 
with uniform dimensions, so that it can divide these into a number of chunks, and provide one $\langle$data, label$\rangle$
chunk pair to each GPU. For different-sized example images, which are the norm in \ac{NHR}, this 
approach breaks down. To fix it, we create a custom DataParallel that accepts the data to come in the form of a list of 
variable-size image tensors. The list is then split into approximately equal-size sub-lists, and a data sub-list with 
corresponding label sub-tensor is provided to each GPU.

\section{Example-Packing}
\label{section:example-packing}

\acp{MDLSTM} parallelization is restricted by the inherent recurrent dependency 
upon previous hidden and memory-states.
In the case of \ac{NHR} based on line-strips, parallelization 
within the computations for a line-strip pixel column is possible. But parallelization 
across pixel columns is not, because of the computational dependencies.
Therefore, it is desirable to exploit the ways of parallelization that are possible to their limit. 
One obvious way to increase parallelization is to increase the batch size.
But this approach, when naively applied, requires all the examples to be 
padded to the same size. For handwriting line-strips or word-strips, which 
are naturally of different dimensions, this approach is computationally 
wasteful in two ways. 
\begin{enumerate}
\item The padding pixels use up a lot of wasted computation. 
\item This  wasted computation coincides with memory waste: 
the space required for the padding could have been used to fit in more real pixels, 
needed for the end result.    
\end{enumerate}

The question is: can we overcome the limitation that all examples need to be of the 
same size, while still respecting the constraints of efficient GPU computation? 
The conclusion is, we can, by using a combination of:

\begin{enumerate}
\item \textbf{Tiling} together examples together to using a greedy space-filling algorithm  to use 
the available space as much as possible.
\item \textbf{Separating pixels} between the tiled examples.
\item \textbf{Binary masking} to block the hidden-state and memory-state input from predecessor cells 
that are not valid input cells but padding cells or separator cells according to the binary mask.    
\end{enumerate}

\begin{figure}[h!]
\begin{center}
  \begin{subfigure}[b]{0.24\textwidth}
  \includegraphics[scale=0.15]{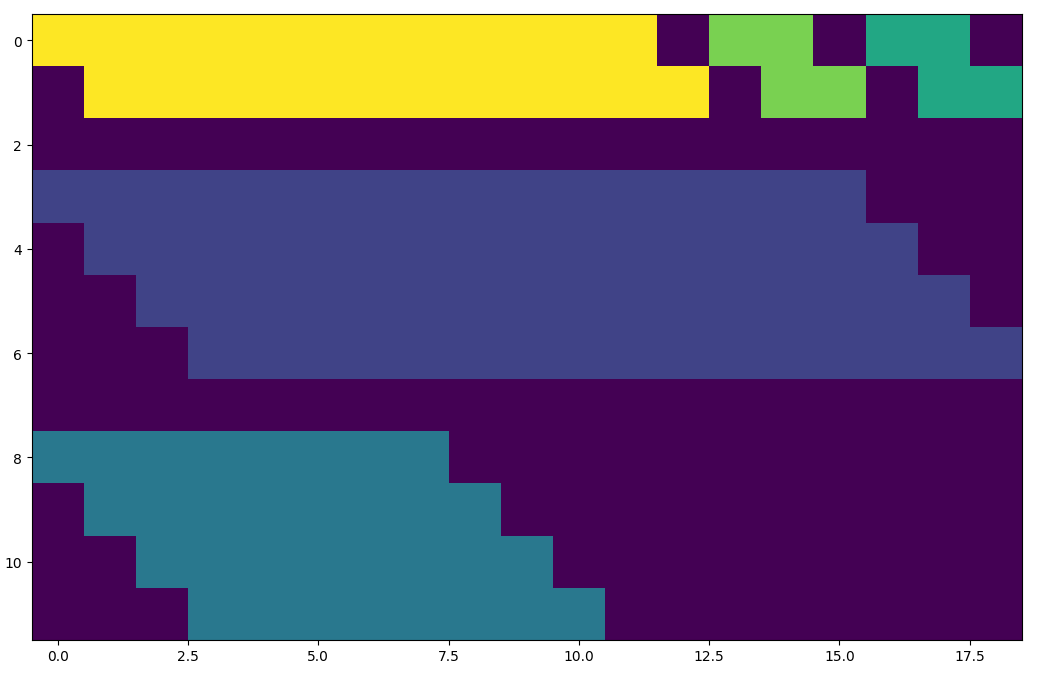}
  \caption{Packed artificial examples}
  \label{figure:example-packing-artificial-example-input}
  \end{subfigure}
  \begin{subfigure}[b]{.24\textwidth}
  \includegraphics[scale=0.30]{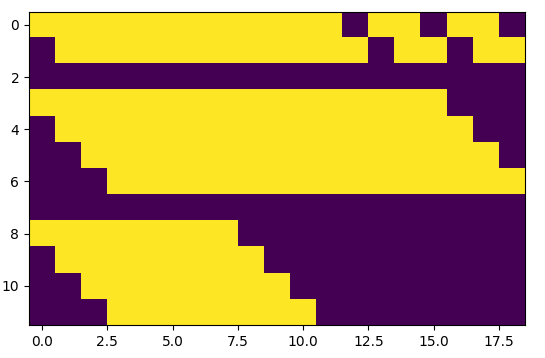}
  \caption{Corresponding binary mask}
  \label{figure:example-packing-artificial-example-mask}
  \end{subfigure}
\end{center}
\caption{Packed artificial data example and corresponding mask.}
\label{figure:example-packing-artificial-example}
\end{figure}

This is best explained by an example. Figure \ref{figure:example-packing-artificial-example-input} shows 
a set of artificial, packed examples, and Figure \ref{figure:example-packing-artificial-example-mask}
the corresponding mask.

\begin{figure*}[h!]
\begin{center}
  \begin{subfigure}[b]{0.3\textwidth}
  \includegraphics[scale=0.3]{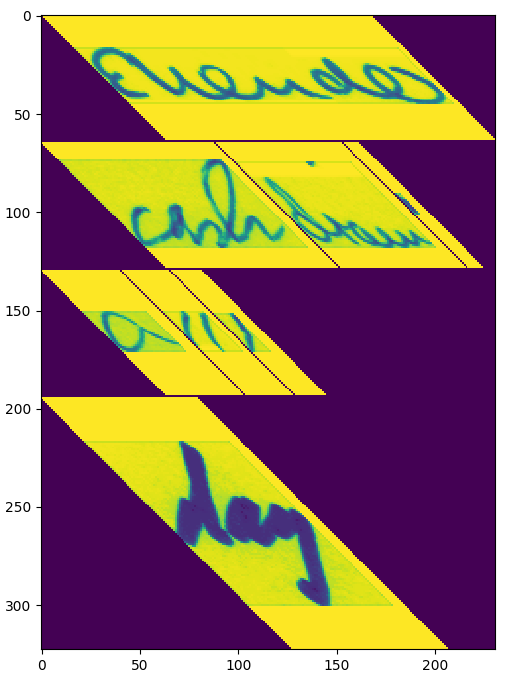}
  \caption{Packed IAM words data example}
  \label{figure:example-packing-iam-words}
  \end{subfigure}
  \begin{subfigure}[b]{0.6\textwidth}
  \includegraphics[scale=0.4]{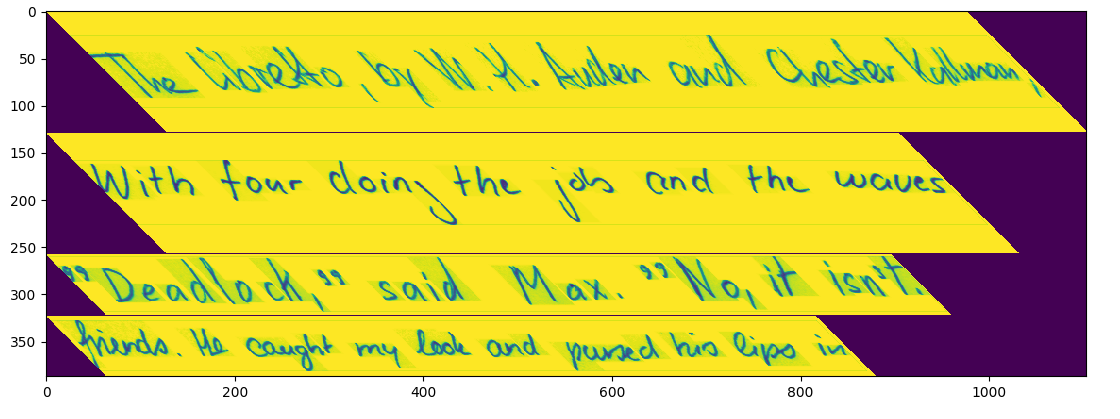} 
  \caption{Packed IAM lines data example}
  \label{figure:example-packing-iam-lines} 
  \end{subfigure}
\end{center}
\caption{Packed IAM words and lines data}
\end{figure*}

Note that examples of the same height are arranged 
in a single row. This is a requirement for allowing column-parallelized \ac{MDLSTM} 
computation with the \emph{input-skewing} trick. 
Furthermore, note that one row of pixels is added between every pair of 
examples. These rows of pixels get skewed diagonally along with the 
input images, as a result of the input-skewing trick. 
Figure \ref{figure:example-packing-iam-words} shows the result of 
example-packing for IAM data examples, consisting of IAM word-strips.
Here, while there is still quite some padding, a large part of it is 
caused by the \emph{input-skewing} trick. But note that while not perfect, 
packing saves out a lot of padding. Without it, every example needs to be padded 
to the largest width and height dimensions occurring in the mini-batch, in this case making
all examples as wide as the first row example and as high as the last row example.
Finally, one may expect that working with line-strips instead of word-strips, 
which for many dataset, such as IAM, is the default, the differences between 
the sizes of the image strips are smaller singe the word-lengths average out.
This is indeed partly true, see Figure \ref{figure:example-packing-iam-lines},
however, at the same time the remaining height-differences imply that 
example-packing can still yield significant savings, predominantly 
in the height dimension.

\subsection{Efficient packing}

With the basic principles behind example-packing explained, the next question is 
how to find a packing that optimally uses the space, tiling the examples in 
a mini-batch such a way to add as little padding as possible. First note that 
there is no need that the dimensions across mini-batches are the same, and this 
fact is exploited in our approach. Second, note the earlier mentioned constraint:
to allow efficient skewing and un-skewing of rows of examples, all examples in the 
row must be of the same height.\EXCLUDE{\footnote{Relaxing this constraint, example (un)skewing 
must be done for every example separately, and a more complicated packing 
algorithm that considers 2-dimensions is required. We leave this for future work.}}
This motivates a four step approach: 1) bucket the examples by height, 2) 
pack the examples in each height bucket, filling up rows, by repeatedely and greedily 
adding the widest still fitting example, 3) tiling the examples first within packed 
rows, then across rows, adding separating pixels in between, 4) applying the input 
skewing trick to create a skewed version of the resulting packed tensor and mask.
In Appendix B we provide pseudocode for the packing algorithm.

Packing is performed just before the computation of each \ac{MDLSTM} layer, based on a list of input tensors for that layer.
After the \ac{MDLSTM} activations are computed on the packed tensors, unpacking is performed on these activations. Figure 
\ref{figure:network-model-structure} indicates these places where packing/unpacking is apllied in the network.
Unpacking is packing in reverse, and consists of the following two steps: 1) the inverse of the input skewing trick is performed to restore the tensor format before skewing, 2) using the original example indices corresponding to the packed examples and their sizes, 
the activations per input example are extracted, while discarding parts of the activations corresponding to separating pixels 
in the input.

\subsection{Packing for block-strided convolution layers}
\label{subsection:packing-for-block-strided-convolution}
Block-strided convolutional layers are convolutional layers with a stride  width and height (``block-size'') corresponding to the 
size of the convolution kernel, such that there is no overlap in input for different  kernel applications. 
These layers layers are used to merge the output of \ac{MDLSTM} layers and decrease resolution. They
require their own pre- and post-processing algorithms to allow efficient processing of input lists obtained from 
\ac{MDLSTM} layers.\footnote{The last fully-connected layer may be considered a special case with a block size of 
1 $\times$ 1.} These two algorithms perform a sort of simplified packing/unpacking.
The \emph{tensor-list chunking} (packing) algorithm chunks a list of input tensors of different sizes into blocks of given size, in our case 
the block-stride of the block-strided convolution layer. The chunking produces for each tensor a list of blocks, and stacks all these blocks 
on the batch simension. This stacked block tensor can be processed very efficiently by a standard 2-D convolution layer.
After computation of the convolutional features, using the size of the tensors in the original input list, a \emph{de-chunking}
algorithm (un-packing) concatenates the output activation blocks again together. 
This application of tensor-list chunking/de-chunking to block-strided convolution is indicated in Figure \ref{figure:network-model-structure} as well. The thus parallely computed result list is equal to what would be 
obtained if the block-strided convolution was computed for each input tensor separately and the result tensors collected in a list.

\section{Effective Optimization} 

We found the use of gradient clipping, particularly the technique proposed in \cite{pmlr-v28-pascanu13}, 
which constitutes rescaling of the gradient to normalize the gradient norm, to be necessary to achieve 
stable learning. Whereas the use of Leaky LP cells was another crucial component, in our experience 
gradient clipping was still needed on top of that to obtain good results. In addition to that, we found 
that the learning rate, max norm for gradient clipping and optimizer needed to be chosen well together.
Unfortunately, this is mostly an empirical matter, in which previous literature can at most help.
We obtained good results, using the Adam optimizer \cite{KingmaEtAl2014} with an initial learning rate of 0.005, and gradient 
clipping using a maximum gradient norm of 10 to be effective. We used the technique of \cite{Denowski2017} to improve Adam, 
by halving the learning rate and resetting the Adam state when the validation scores (WER and CER) got worse, resuming 
training from the best last model. 
Following \cite{Voigtlaender2016}, we trained for a maximum of 80 epochs, 
after which we selected the best performing model on the validation-set, and used this model 
to evaluate on the test-set.

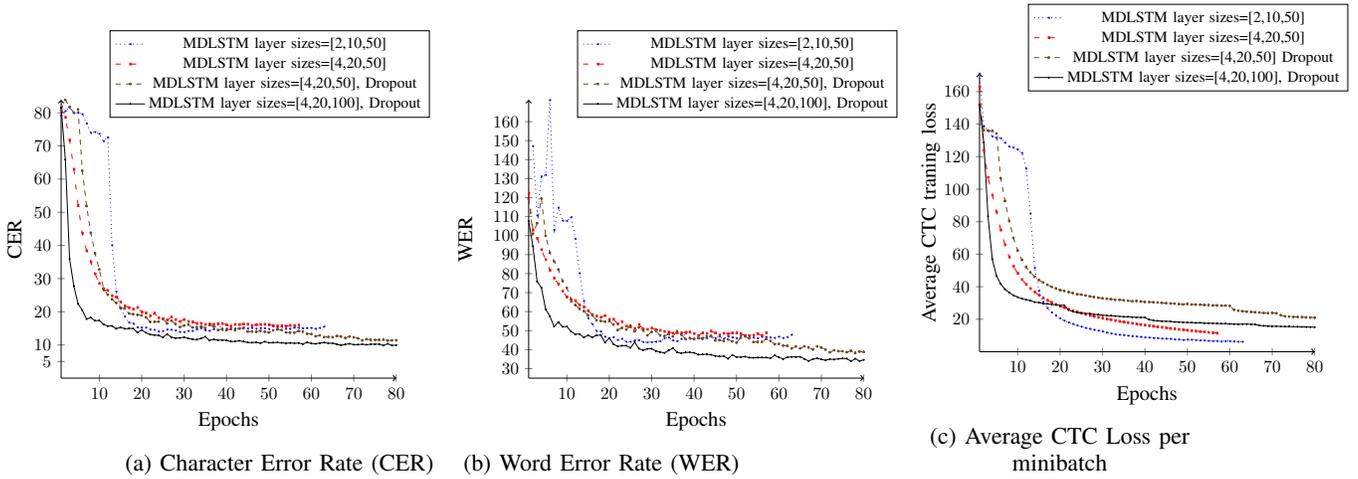
\begin{figure*}
\captionsetup[subfigure]{justification=centering}
\centering
\hspace{-1.0cm}
\begin{subfigure}[b]{.26\linewidth}
\centering
\hspace{-1.9cm}  
\scalebox{0.65}{
\begin{tikzpicture}[every mark/.append style={mark size=0.5pt}]
\begin{axis}[
legend style={font=\fontsize{9}{9}\selectfont, at={(1.1,1.25)}},
ymin=0,
xtick={10,20,30,40,50,60,70,80},
ytick={5,10,20,30,40,50,60,70,80},
axis lines=middle,
axis line style={->},
x label style={at={(axis description cs:0.5,-0.1)},anchor=north, font=\large},
y label style={at={(axis description cs:-0.1,.5)},rotate=90,anchor=south, font=\large},
xlabel={Epochs},
ylabel={CER}]
\addplot+[line width=0.5pt, dotted] table [x=Epoch, y=CER_mdlstm_layers_2-10-50, col sep=comma, mark options={scale=10}] {iam-scores-dropout-initial-learning-rate-0.005-mdlstm-sizes-4-20-100-CER.csv};]
\addlegendentry{MDLSTM layer sizes=[2,10,50]}
\addplot+[line width=0.5pt, loosely dashed] table [x=Epoch, y=CER_mdlstm_layers_4-20-50, col sep=comma, mark options={scale=10}] {iam-scores-dropout-initial-learning-rate-0.005-mdlstm-sizes-4-20-100-CER.csv};]
\addlegendentry{MDLSTM layer sizes=[4,20,50]}
\addplot+[line width=0.5pt, dashed] table [x=Epoch, y=CER_last_mdlstm_layer_50_DROPOUT, col sep=comma, mark options={scale=10}] {iam-scores-dropout-initial-learning-rate-0.005-mdlstm-sizes-4-20-100-CER.csv};]
\addlegendentry{MDLSTM layer sizes=[4,20,50], Dropout}
\addplot+[line width=0.5pt] table [x=Epoch, y=CER_last_mdlstm_layer_100_DROPOUT, col sep=comma, mark options={scale=10}] {iam-scores-dropout-initial-learning-rate-0.005-mdlstm-sizes-4-20-100-CER.csv};]
\addlegendentry{MDLSTM layer sizes=[4,20,100], Dropout}
\end{axis}
\end{tikzpicture}}
\caption{Character Error Rate (CER)}\label{figure:CERIAM}
\end{subfigure}
\hspace{-0.5cm}
\begin{subfigure}[b]{.24\linewidth}
\centering
\raisebox{0.0cm}{
\scalebox{0.65}{
\begin{tikzpicture}[every mark/.append style={mark size=0.5pt}]
\begin{axis}[
legend style={font=\fontsize{9}{9}\selectfont, at={(1.1,1.25)}},
ymin=25,
xtick={10,20,30,40,50,60,70,80},
ytick={30,40,50,60,70,80,90,100,110,120, 130, 140, 150, 160},
axis lines=middle,
axis line style={->},
x label style={at={(axis description cs:0.5,-0.1)},anchor=north, font=\large},
y label style={at={(axis description cs:-0.15,.5)},rotate=90,anchor=south, font=\large},
xlabel={Epochs},
ylabel={WER}]
\addplot+[line width=0.5pt, dotted] table [x=Epoch, y=WER_mdlstm_layers_2-10-50,  col sep=comma, mark options={scale=10}] {iam-scores-dropout-initial-learning-rate-0.005-mdlstm-sizes-4-20-100-WER.csv};]
\addlegendentry{MDLSTM layer sizes=[2,10,50]}
\addplot+[line width=0.5pt, loosely dashed] table [x=Epoch, y=WER_mdlstm_layers_4-20-50,  col sep=comma, mark options={scale=10}] {iam-scores-dropout-initial-learning-rate-0.005-mdlstm-sizes-4-20-100-WER.csv};]
\addlegendentry{MDLSTM layer sizes=[4,20,50]}
\addplot+[line width=0.5pt, dashed] table [x=Epoch, y=WER_last_mdlstm_layer_50_DROPOUT,  col sep=comma, mark options={scale=10}] {iam-scores-dropout-initial-learning-rate-0.005-mdlstm-sizes-4-20-100-WER.csv};]
\addlegendentry{MDLSTM layer sizes=[4,20,50], Dropout}
\addplot+[line width=0.5pt] table [x=Epoch, y=WER_last_mdlstm_layer_100_DROPOUT, col sep=comma, mark options={scale=10}] {iam-scores-dropout-initial-learning-rate-0.005-mdlstm-sizes-4-20-100-WER.csv};]
\addlegendentry{MDLSTM layer sizes=[4,20,100], Dropout}
\end{axis}
\end{tikzpicture}}}
\caption{Word Error Rate (WER)}\label{figure:WERIAM}
\end{subfigure}%
\hspace{1.8cm}
\begin{subfigure}[b]{.22\linewidth}
\centering
\hspace{-1.5cm}  
\scalebox{0.65}{
\begin{tikzpicture}[every mark/.append style={mark size=0.5pt}]
\begin{axis}[
legend style={font=\fontsize{9}{9}\selectfont, at={(1.1,1.25)}},
ymin=0,
xtick={10,20,30,40,50,60,70,80},
ytick={20,40,60,80,100,120,140,160},
axis lines=middle,
axis line style={->},
x label style={at={(axis description cs:0.5,-0.1)},anchor=north,font=\large},
y label style={at={(axis description cs:-0.1,.5)},rotate=90,anchor=south,font=\large},
xlabel={Epochs},
ylabel={Average CTC traning loss}]
\addplot+[line width=0.5pt, dotted] table [x=Epoch, y=CTC-LOSS_mdlstm_layers_2-10-50,  col sep=comma, mark options={scale=10}] {iam-scores-dropout-initial-learning-rate-0.005-mdlstm-sizes-4-20-100-CTC-LOSS.csv};]
\addlegendentry{MDLSTM layer sizes=[2,10,50]}
\addplot+[line width=0.5pt, loosely dashed] table [x=Epoch, y=CTC-LOSS_mdlstm_layers_4-20-50,  col sep=comma, mark options={scale=10}] {iam-scores-dropout-initial-learning-rate-0.005-mdlstm-sizes-4-20-100-CTC-LOSS.csv};]
\addlegendentry{MDLSTM layer sizes=[4,20,50]}
\addplot+[line width=0.5pt, dashed] table [x=Epoch, y=CTC-LOSS_last_mdlstm_layer_50_DROPOUT,  col sep=comma, mark options={scale=10}] {iam-scores-dropout-initial-learning-rate-0.005-mdlstm-sizes-4-20-100-CTC-LOSS.csv};]
\addlegendentry{MDLSTM layer sizes=[4,20,50] Dropout}
\addplot+[line width=0.5pt] table         [x=Epoch, y=CTC-LOSS_last_mdlstm_layer_100_DROPOUT, col sep=comma, mark options={scale=10}] {iam-scores-dropout-initial-learning-rate-0.005-mdlstm-sizes-4-20-100-CTC-LOSS.csv};]
\addlegendentry{MDLSTM layer sizes=[4,20,100], Dropout}
\end{axis}
\end{tikzpicture}}
\caption{Average CTC Loss per minibatch}\label{figure:AverageCTCLossIAM}
\end{subfigure}%
\caption{Results on the IAM validation set for MDLSTM \ac{NHR} networks trained with or without dropout. 
For these results, decoding is done using a beam-search decoder without language model.}
\label{figure:ResultGraphsIAM}
\end{figure*}

\section{Experiments}

\subsection{Dataset}
\label{subsection:dataset}

We perform experiments on the IAM-database \cite{IAM-Database}, which is an English multi-writer 
handwriting dataset based on material from the Lancaster-Oslo/Bergen (LOB) corpus. We chose this 
dataset as it is one of the most frequently used benchmark datasets in the field, and a such 
facilitates easy comparison to other works including \cite{PhamEtAl2014, Voigtlaender2016, Puigcerver2017}. 
As such, during our reimplementation of \ac{MDLSTM}-based \ac{NHR} models from scratch,
the dataset was invaluable for testing where we stood with our system in comparison to earlier work.
The IAM database is of moderate size. It contains material of 657 different writers, 
and is partitioned into subsets for validation, training and testing of 161, 966 and 2 915 lines.
This data split corresponds to the split of the IAM (lines) dataset used in \cite{PhamEtAl2014}, \cite{Voigtlaender2016} and \cite{Puigcerver2017}. For the IAM words dataset we used the same split files, and obtained subsets for  training, validation and testing of 
55079, 8895 and 25920 words. Unfortunately, somehow these sizes for the words dataset do not match the word-set sizes for 
training, validation and testing  reported in  \cite{PhamEtAl2014} (80421, 16770, 17991)
even though they were derived from the same data spit files. This makes an exact quality comparison with \cite{PhamEtAl2014} for the word recognition systems not possible.\footnote{We took the data splits from Th\'{e}odore Bluche, as available from his website http://www.tbluche.com/resources.html. This yielded matching sizes for IAM lines, but for some reason not for IAM words. }
However, since our main quality comparison is on line recognition, this is not a major problem, 
and we leave further investigation of this issue for future work.
In combination with the IAM dataset, we use a domain-specific language model trained on material from the 
(unused parts of the) LOB corpus and the Brown corpus.

\subsection{Results}
Figure \ref{figure:ResultGraphsIAM} shows graphs tracking the model performance across training progress, in addition 
Table \ref{fig:results_on_iam_test-set_using_language_model} shows the results on the validation and test-set, using the 
best performing validation model and applying it to the test-set. Figure \ref{figure:CERIAM}
and \ref{figure:WERIAM} show the \ac{CER} and \ac{WER} scores on the validation set, without use of a language model during decoding; and 
Figure \ref{figure:AverageCTCLossIAM} shows the average CTC loss per epoch. 
From these graphs, and the table, a few observations can be made. First, the models trained using dropout
outperform these without dropout. Second, whereas in case dropout is used the larger model with \ac{MDLSTM} layer sizes of  4, 20 and 100
performs best, without dropout this model performs worse than the smaller model with \ac{MDLSTM} layer sizes of 2, 10 and 50. This is in 
line with what has been reported earlier in the literature, i.e. in \cite{PhamEtAl2014}. Third, the models without dropout 
have a CTC loss graph that keeps going down, whereas the \ac{CER} and \ac{WER} start to increase (i.e. worsen) again after a certain number 
of epochs, indicating over-fitting. In contrast, the CTC loss graphs for the models that use dropout flatten out faster after some point,
whereas the quality of the models as measured by \ac{CER} and \ac{WER} for these models keeps increasing nearly till the end.
Last, both systems with dropout not only give final better results, but also improve faster than the models without dropout, with the largest 
model with dropout showing a markedly faster improvement towards a decent system than all other systems. 
In summary, these graphs show that dropout is highly effective both in delivering superior results, as well as in our case in 
delivering them faster.

Table \ref{fig:results_on_iam_comparison_to_literature} shows a comparison of our best models against the results by \cite{PhamEtAl2014} and
\cite{Voigtlaender2016} in the literature. Comparing to the results of \cite{PhamEtAl2014}, it can be noted that our best system using dropout
is still slightly outperformed by theirs, but the difference is small. The remaining difference might still be explained by the fact that 
they use a form of curriculum learning, training first on the words and then on the lines, as well as using a different optimizer.

We mostly followed \cite{Voigtlaender2016} in our training approach, not using curriculum learning and using Adam instead of SGD (or RMSPRob)
as an optimizer. The lower performance we obtain in comparison to \cite{Voigtlaender2016} is explainable from the our different network structure, 
which is chosen almost identical as \cite{PhamEtAl2014}. 
As our work focusses on proposing computational gains, which apply also for even fancier networks, 
for example adding max-pooling layers and layer normalization, we consider the fact that in terms of recognition accuracy our model performs slightly 
below state-of the-art not to be a big problem.

Table \ref{fig:results_on_iam_words} shows our results on IAM words and the reported results on IAM words from \cite{PhamEtAl2014}. Our results on 
IAM words without vocabulary are worse than those of \cite{PhamEtAl2014} in this setting. However, as discussed before, these scores cannot really be compared. 
Using the same dataset split as for IAM lines,  our training set became considerably smaller than the size reported in \cite{PhamEtAl2014}, 
which probably explains the score-differences.

\begin{table}
\caption{Recognition quality results on the IAM  lines validation-set and test-set.}
\centering
 \begin{tabular}{|l|l|l|l|l|}
 \hline
 & \multicolumn{2}{|c|}{validation} &  \multicolumn{2}{|c|}{test}\\
 \hline
 System & WER & CER & WER & CER \\
 \hline
 Leaky LP Cell [2,10,50], no dropout  & 43.8 & 13.9 & 52.2 & 18.8  \\
 \ \ \ \ \ \ \ \ + Vocabulary and LM  & 15.6  & 6.4 & 22.1 &  9.9 \\
 Leaky LP Cell [4,20,50], no dropout  & 47.4 & 15.7  & 54.6  &  20.4  \\
 \ \ \ \ \ \ \ \ + Vocabulary and LM  &  19.1  & 8.9  & 25.1 &  12.9  \\
 \hline
 Leaky LP Cell [4,20,50], dropout  & 38.4 & 11.3 & 44.0 & 14.5 \\
 \ \ \ \ \ \ \ \ + Vocabulary and LM  & 17.7 &  7.8 & 18.6 &  8.3 \\
 Leaky LP Cell [4,20,100], dropout  & 33.9 & 9.8& 40.8 & 12.9  \\
 \ \ \ \ \ \ \ \ + Vocabulary and LM  & 15.5 & 6.4 & 15.9 & 6.6 \\
 \hline
 \end{tabular}
\label{fig:results_on_iam_test-set_using_language_model}
\end{table}

\begin{table}
\caption{Comparison to literature results on IAM lines.}
\centering
 \begin{tabular}{|l|l|l|l|l|}
 \hline
 & \multicolumn{2}{|c|}{validation} &  \multicolumn{2}{|c|}{test}\\
 \hline
 System & WER & CER & WER & CER \\
 \hline
 Leaky LP Cell [4,20,100], dropout  & 33.9 & 9.8& 40.8 & 12.9  \\
 \ \ \ \ \ \ \ \ + Vocabulary and LM  & 15.5 & 6.4 & 15.9 & 6.6 \\
 \hline 
 \hline 
 Pham et.al (2014), no dropout & 36.5 & 10.4 &  43.9 & 14.4 \\
 \ \ \ \ \ \ \ \ + Vocabulary and LM & 12.1 & 4.2 &  15.9 & 6.3 \\
 Pham et.al (2014), dropout & 27.3 & 7.4 &  35.1 & 10.8  \\
 \ \ \ \ \ \ \ \ +  Vocabulary and LM & 11.2 & 3.7 &  13.6 & 5.1  \\
 Voigtlaender et.al (2016) & 7.1 & 2.4 & 9.3 & 3.5  \\
 \hline
 \end{tabular}
 \label{fig:results_on_iam_comparison_to_literature}
\end{table}

\begin{table}
\caption{Recognition quality for IAM words and literature results using a larger training set (see section \ref{subsection:dataset}). }
\centering
 \begin{tabular}{|p{3.9cm}|l|l|l|l|}
 \hline
 & \multicolumn{2}{|c|}{validation} &  \multicolumn{2}{|c|}{test}\\
 \hline
 System & WER & CER & WER & CER \\
 \hline
 Leaky LP Cell [4,20,100], dropout  & 33.58 & 14.11 & 42.05 & 19.15  \\
 \ \ \ \ \ \ \ \ + Vocabulary  & 20.12 & 10.42 & 25.66 & 14.29 \\
 \hline 
 \hline 
 Pham et.al (2014), dropout &  --- & --- & 31.44 & 14.02  \\
 \hline
 \end{tabular}
 \label{fig:results_on_iam_words}
\end{table}

\begin{table}
\caption{Memory and time usage for models with and without example-packing, with batch sizes chosen the maximal possible 
given the observed maximum GPU memory usage.} 
\begin{tabular}{|p{2.0cm}| p{0.4cm} |p{1cm}|p{0.9cm}|p{1.1cm}|p{1.1cm}|}
\hline
Preparation of batch examples & batch size & time per epoch (HH:MM: SS)  & examples per second & max GPU1 memory use (MB)   & max GPU2 memory use (MB) \\
\hline
\multicolumn{6}{|c|}{IAM lines} \\
\hline
batch-padding & 8 &  07:24:06 & 0.243 & 10824  & 10675  \\   
\hline
example-packing & 12 &   05:04:45 &  0.355  & 10694  & 10780  \\    
\hline
\multicolumn{6}{|c|}{IAM words} \\
\hline
batch-padding & 20 &  06:26:48 & 2.38 & 11074 & 11144 \\
\hline
example-packing & 200 &   00:58:22. &  16.1  & 10827  & 10849  \\  
\hline
\end{tabular}

\label{table:memory-and-time-usage-with-or-without-packing}
\end{table}

\subsection{Impact of packing on the training time}

In this section we show the impact of packing on the training times in both the (IAM) line-recognition and 
(IAM) word-recognition scenario. In the setting were packing is not used, we instead use a strategy we will call 
\ac{LMBR}, padding examples for each batch (on-the-fly) to the maximum height and width occurring within that batch. 
\ac{LMBR} is already a faster baseline than padding all examples within the training set to the same maximum height and with, which is a simple 
but computationally wasteful strategy.

Table \ref{table:memory-and-time-usage-with-or-without-packing} shows the time consumed 
per epoch, examples per second and maximum GPU memory usages for identical models that 
were trained with or without example-packing. For these experiments, we used our best performing
model with MDLSTM layers of sizes 4, 20 and 100 plus dropout, while the batch 
sizes were chosen to be maximal given the peak GPU memory consumption for the setting. This yielded 
for line recognition batch sizes of 8 when no packing was used, and 12 when it was used and 
for word recognition batch sizes of 20 when no packing was used, and 200 when it was used 
As can be seen, these settings yield to similar maximum memory usage, and neither of the batch sizes could
be further increased without running out of the total available GPU memory (11178 MB).\footnote{In case of IAM words, we approximated the maximum batch size not giving out of memory problems by trial with a step size of 5. 
Given the large difference in the maximum possible batch sizes with and without packing in this setting, this level of precision is adequate for the purpose of our comparison, 
and finding the exact maximal possible batch size would not significantly change the results.}

For line recognition, looking at the times per epoch or  examples per second, it can be observed that using packing the same 
computation can be done in 69\% of the time used without packing. 
When testing on line strips, the savings come mostly from avoiding the need to pad 
al examples within a batch to the same height, as in this case the width differences between words average out to 
a large extent. Even so, already in this settings packing makes a noticeable difference.
Looking at word recognition next, the speed improvements are more drastic. Whereas without packing, one epochs takes about six and a half hours, 
with packing it takes only about an hour, thanks to the major gains in efficiency packing yields in this setting, indicated by the 
much larger possible batch size of 200. This major speedup, by a factor 6.6, is even relevant for models which use convolutional layers to 
replace MDLSTMs \cite{Puigcerver2017}. Since while these systems are perhaps more efficient to begin with, they still waste a lot of computation on padding 
and could therefore benefit from packing, with some adaptations for the changed network structure.

\textit{Are variable batch sizes an alternative to packing?}\\
Whereas in case of word-recognition, without packing on average about 75\% of the input pixels consists of padding, this factor four saving does not 
explain the even larger difference in possible batch sizes. The reason that the maximum batch size using packing (200) is ten times larger rather than ``just'' four times 
larger than the size without packing (20), is that for the maximum possible batch size the ``worst-case'' batch counts, and not the average batch. 
That is, a single batch with one very high and one very wide example creates a peak in memory usage, and the batch size must be chosen to allow this peak value 
to still fit in the maximal GPU memory. In contrast, when packing is used (and padding mostly avoided), the fluctuation in effective input size and consequently GPU
memory usage is much smaller.
This suggests that as a partial alternative to packing, some savings could also be made by using a variable batch size, which resizes 
based on the size of the examples within the batch.
However, this gives its own complications in terms of efficient data loading. It may also require additional measures to be taken in order to keep stable learning. Finally, note that saving out most of the factor-four blowup in size because of padding using example-packing is by itself substantial, and this saving can only be realized by packing, 
not by using variable batch sizes.

\section{Discussion}
While the packing techniques as applied in this paper are specific to \acp{MDLSTM}, the general principle 
of efficient packing and unpacking of variable-size examples provided as a list is general enough to be adapted for large speed 
improvements of many deep learning models that work with variable-sized inputs. Notably, the tensor-list chucking algorithm as discussed in 
section \ref{subsection:packing-for-block-strided-convolution} is itself an illustration of how the idea first developed for \ac{MDLSTM}
layers was then adapted for convolution layers with non-overlapping strides as well. Whereas the generalization of this algorithm to 
general convolution layers is slightly more complex than the algorithm described here, it is of a similar form.
In future work we would like to further generalize the principle of packing, so that more different network types dealing with variable-size 
inputs can benefit from it.

\section{Conclusion} 

We presented several new methods that can help to drastically increase the speed of deep learning models using \acp{MDLSTM}. 
One of these techniques, \emph{example-packing}, achieved a factor 6.6 speed improvement on word-based handwriting-recognition,
by avoiding wasted computation on padding. Our methods were thoroughly tested on a \ac{MDLSTM}-based implementation of 
state-of-the-art neural handwriting recognition models implemented with PyTorch.

\section*{Acknowledgment}  

This research has been supported by the ADAPT Centre for Digital Content Technology which is funded under the SFI Research Centres Programme (Grant 13/RC/2106) and is co-funded under the European Regional Development Fund.
\noindent 
\includegraphics[width=1cm]{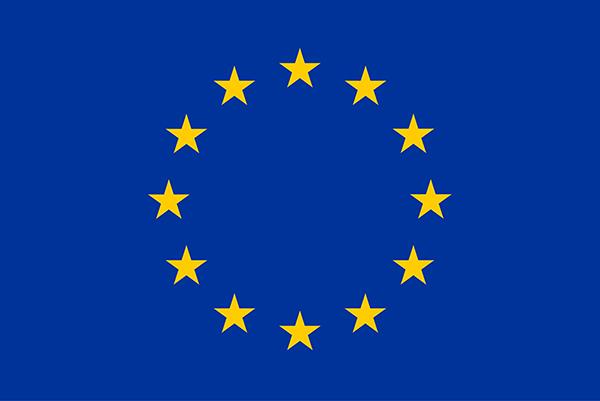}
This work has also received funding from the European Union's Horizon 2020 research and innovation programme under the Marie Sk{\l}odowska-Curie grant agreement No 713567.
Many thanks to Joost Bastings for invaluable consultation on deep learning technology and best practices. Special thanks also to Paul Voigtlaender and 
Th\'{e}odore Bluche for their helpful and generous advise which has been important in getting MDLSTMs for handwriting recognition (from scratch) to work.

\bibliographystyle{IEEEtran}
\bibliography{references}

\clearpage

\onecolumn

\section*{Appendix}

\subsection{Further application of convolutions with grouping}
\label{appendix:application-convolution-with-grouping}

For computing fully-connected layers that take input from hidden states  and memory states, convolutions with grouping 
are applied to increase parallelization. There are five matrix computations necessary 
for each of the two hidden states $H_1$ and $H_2$, and three for each of the memory states $S_1$ and $S_2$. 
Since the hidden- and memory state computations have the same input dimensionality, 
we can all compute them using a single convolution with grouping, using replication of the input to overcome the fact that 
the number of output groups differs for the hidden- and memory states. 
Finally, we can shift the output by once cell for the second hidden- and memory state respectively, 
to get a vector of activations of the top ancestor states but overlayed with the activations of the \emph{left} ancestor states. 
That way the activation tensors for the two hidden/memory states can be directly summed to combine them for further computation.

\subsection{Packing Algorithm Details}
\label{appendix:packing-algorithm-pseudocode}

\begin{algorithm*}
 \KwData{List of examples}
 \KwResult{$t_{packed\_examples}$, $t_{masks}$: tensors, $original\_example\_indices$ :
  2-D array datastructure storing the original example indices for the examples in the filled rows}
  $height\_buckets$ := Bucket the examples into groups of the same height ;\\
  $packing\_rows$ :=  [] ; $original\_example\_indices$ = [] ; $current\_row$ := [] ; $current\_indices\_row$ := [] ; \\
  \For{height\_bucket $\in$ height\_buckets}{
  \While {not\_empty(height\_bucket)}{
    $example$, $original\_example\_index$ := largest\_fitting\_example($current\_row$, $height\_bucket$) ; \\
    \uIf{example != None}{      
      $current\_row$.append(example) ; current\_indices\_row.append(original\_example\_index) ; \\
   }
   \Else{
    \# Current row is full, start a new row \\
    $packing\_rows$.append(current\_row) ; $original\_example\_indices$.append(current\_indices\_row) ; \\
    $current\_row$ = [] ; current\_indices\_row = [] ;
   }
  }
  }
  $t_{packed\_examples}$ := concatenate\_packing\_rows\_adding\_separation\_pixels($packing\_rows$) ;\\
  $t_{masks}$ := create\_mask($t_{packed\_examples}$) ; \# all example indices replaced by 1, and all separator pixels replaced by 0   \\
  $t_{packed\_examples\_skewed}$ := apply\_input\_skewing($t_{packed\_examples}$) ; $t_{masks\_skewed}$ := apply\_input\_skewing($t_{masks}$) ; \\
  \Return $t_{packed\_examples\_skewed}$, $t_{masks\_skewed}$, $original\_example\_indices$
  \vspace{0.2cm}
 \caption{Packing Algorithm}
 \label{algorithm:packing_algorithm}
\end{algorithm*}

\subsection{Experimental details and source-code}
\label{appendix:code}
All our experiments were done using PyTorch. We used two NVIDIA GEFORCE\textsuperscript{\textregistered} GTX 1080 graphic cards to run our experiments.
In our work, for implementing ctc-loss, we used PyTorch bindings for warp-ctc \cite{deep-speech2016} by Baidu research.\footnote{https://github.com/baidu-research/warp-ctc}
Our version of the code\footnote{https://github.com/gwenniger/warp-ctc.}, was forked and slightly adapted from the one by Sean Naren and others\footnote{https://github.com/SeanNaren/warp-ctc}.
Additionally we used a slightly adapted version\footnote{https://github.com/gwenniger/ctcdecode.} of the ctc beam-search decoder for PyTorch, 
developed by Ryan Leary and others\footnote{https://github.com/parlance/ctcdecode}.
This decoder supports KenLM \cite{Heafield-kenlm} word-based n-gram
language models, which we use in our experiments. The rest of our source-code is intended to be made available together with a peer-reviewed publication of the work.

\end{document}